%% file: ms.tex
\documentclass[12pt]{article}    

\input{preamble.tex}

\title{Robustness of Control Design via Bayesian Learning} 

\author{Nardos Ayele Ashenafi \\
        Electrical and Computer Engineering Department \\
        Boise State University \\
        \And
        Wankun Sirichotiyakul \\
        Electrical and Computer Engineering Department \\
        Boise State University \\
        \And
        Aykut C. Satici \\
        Mechanical and Biomedical Engineering Department \\
        Boise State University \\
}

\begin{document}

\maketitle    
\input{contents/abstract.tex}



\renewcommand{\arraystretch}{1.15}
\input{contents/introduction.tex}

\input{contents/justification.tex}

\input{contents/conclusion.tex}

\bibliographystyle{ieeetr}        
\bibliography{ms.bib}           


\end{document}

%% file: preamble.tex
\usepackage{PRIMEarxiv}

\usepackage[utf8]{inputenc} 
\usepackage[T1]{fontenc}    

\usepackage{amsmath}
\usepackage{amssymb}
\usepackage{amsfonts}
\usepackage{amsfonts}       
\usepackage{booktabs}       
\usepackage{cancel}
\usepackage{hyperref}       
\usepackage{lipsum}
\usepackage{fancyhdr}       
\usepackage{graphics}
\usepackage{graphicx}
\graphicspath{{figures/}}

\usepackage{mathtools}
\usepackage{microtype}      
\usepackage{nicefrac}       
\usepackage{url}            
\usepackage[table,x11names,svgnames,dvipsnames]{xcolor}
\usepackage{todonotes}
\usepackage{wrapfig}

\makeatletter

\newcommand{\Rmnum}[1]{\expandafter\@slowromancap\romannumeral #1@}
\makeatother

\makeatletter
\newcommand{\pushright}[1]{\ifmeasuring@#1\else\omit\hfill$\displaystyle#1$\fi\ignorespaces}
\newcommand{\pushleft}[1]{\ifmeasuring@#1\else\omit$\displaystyle#1$\hfill\fi\ignorespaces}
\makeatother

\newcommand{\abs}[1]{\left|{#1}\right|}

\newcommand{\mc}[1]{\mathcal{#1}}

\newtheorem{pf}{Proof}
\newtheorem{lem}{Lemma}

\makeatletter
\let\NAT@parse\undefined
\makeatother
\usepackage{hyperref}
\hypersetup{
    unicode=false,          
    pdftoolbar=true,        
    pdfmenubar=true,        
    pdffitwindow=false,     
    pdfstartview={FitH},    
    pdftitle={Robustness of Control Design via Bayesian Learning},    
    pdfauthor={Nardos Ayele Ashenafi, Wankun Sirichotiyakul, Aykut C. Satici},     
    pdfnewwindow=true,      
    colorlinks=true,       
    linkcolor=magenta,          
    linkbordercolor=orange,
    citecolor=blue,        
    citebordercolor=green,
    filecolor=magenta,      
    urlcolor=cyan,           
    urlbordercolor=blue,
}

%% file: contents/abstract.tex
\begin{abstract} 

In the realm of supervised learning, Bayesian learning has shown robust
predictive capabilities under input and parameter perturbations.
Inspired by these findings, we demonstrate the robustness properties of Bayesian 
learning in the control search task.
%
%
We seek to find a linear controller that stabilizes a one-dimensional open-loop
unstable stochastic system. We compare two methods to deduce the controller: the
first (deterministic) one assumes perfect knowledge of system parameter and
state, the second takes into account uncertainties in both and employs Bayesian
learning to compute a posterior distribution for the controller.
%
%
%

\end{abstract}

%% file: contents/introduction.tex
\section{Introduction}


%
%
%
%
In model-based control theory, nominal system models are used to rigorously
generate optimal policies and provide stability guarantees.
In the presence of model uncertainties and measurement noise, model-based
controllers may show poor performance or even instability.
Data-driven control design
techniques~\cite{heess2017emergence,lillicrap2015continuous} address this issue
by taking away the need for a system model and directly learning viable policies
through repeated interactions with the real system. 
Nevertheless, the stability and robustness analysis under these techniques are
limited to the visited state space.
Moreover, to avoid hardware damages, training may require conservative state
constraints, consequently limiting the capabilities of the learning
technique~\cite{ARGALL2009469, beaudoin2021structured}.

Bayesian learning (BL) offers a way to simultaneously utilize the learning
framework of data-driven techniques and address the uncertainties in
model-based techniques.
A common approach is Bayesian model learning, which infers a stochastic
dynamical model from a series of interactions with the real system, from which a
controller is designed~\cite{sadigh2015safe, shen2022online,
pmlr-v54-linderman17a}. 
Other frameworks take this approach one step further, by using BL to
simultaneously quantify model uncertainties and learn viable policies while
imposing Lyapunov stability constraints~\cite{fan2020bayesian}.
Such techniques have become common with safety critical systems, such as space
exploration vehicles and drones, that need to interact and make decisions in an
uncertain environment with low risk of damage.

The robustness properties of BL are also unmatched;
%
%
it provides a principled way of adding prior information to learn probability
distributions over the parameters of the model, while taking into account the
uncertainty in the learning process~\cite{murphy2012machine}.
The results in~\cite{cardelli2019robustness} and~\cite{cardelli2019statistical}
have shown that in contrast to deterministic models, whose parameters are given
by point estimates, Bayesian models are more robust against input and parameter
perturbations.
Moreover, Bayesian adversarial learning techniques intentionally add
uncertainties to consider the potential effects of adversaries in the learning
process~\cite{wicker2021bayesian}. Such techniques have shown superiority over
point estimates in adversary accommodation and robust
performance~\cite{ye2018bayesian}.

In this work, we bypass the need to learn a stochastic dynamical model and
directly employ BL on the control search task. 
The goal is to learn robust controllers under system parameter and
measurement uncertainties.
%
%
%
We provide a comparison study on the robustness properties of deterministic and
Bayesian solutions to the optimal control problem.
The study examines error sensitivity of the deterministic solution, which
inherently assumes perfect knowledge of the system parameters and measured
states.
We also explore the advantages of reasoning about system parameter and
measurement uncertainties into learning a stochastic optimal controller via
Bayesian learning technique.

%% file: contents/justification.tex
\section{Theoretical Justification for the Robustness Properties of Bayesian Learning}
\label{sec:justification}

In this section, we demonstrate the improved robustness properties of Bayesian
learning over point-estimates of a policy. 
%
%
We start with an open-loop unstable $1$-dimensional affine control system, over
which we close the loop with a linear controller whose sole parameter is to be
determined in an optimal manner. We solve and analyze two stochastic optimal
control problems: (i) optimal control of the nominal system subject to only
parameter uncertainty, where the only system parameter $p$ is distributed
according to a Gaussian distribution, (ii) optimal control of the nominal system
subject to both parameter and measurement uncertainties, where the control
signal is corrupted due to imperfect measurement of the state $x$. 

We compare the optimal controller found by assuming perfect knowledge of the
system parameter to one that finds a posterior probability distribution that
takes into account system uncertainties. We show that as the uncertainty grows,
the two solutions may be quite far apart, and that marginalizing the probability
distribution over the posterior distribution produces more robust controllers;
that is, controllers that stabilize a wider range of system parameters. Our
results also show that the risk of instability increases drastically as the
measurement error increases, further favoring the use of Bayesian learning to
infer a posterior distribution before making controller decisions.

\subsection{Optimal Control under Parameter Uncertainty}
Let us consider the first-order scalar control system, whose system parameter
$p$ is uncertain: 
\begin{equation} 
    \begin{cases} 
        \dot{x} = px + u, \; x(0) = x_0, \\ 
        u(x) = \theta x.  
    \end{cases} 
    \label{eq:first-order} 
\end{equation}
We assume that $p \sim \mc{N}\left(\hat{p}, \sigma_p^2\right)$ where $\hat{p}$
designates our best prior point estimate of the system parameter $p$ and
$\sigma_p > 0$ quantifies the uncertainty in the knowledge of the system
parameter. The controller is set to be linear in the state $x \in \mathbb{R}$
with its only parameter $\theta \in \mathbb{R}$ to be determined through
optimization. Without loss of generality, we will take the initial
condition $x_0 = 1$. The performance index to be optimized for determining the
best control parameter $\theta$ is
\begin{equation} \mc{J} = \int_0^T \left(\frac{1}{2}qx^2 + \frac{1}{2}ru^2 \right) d t,
\label{eq:perfind} \end{equation}
where $T$ is the control horizon and $q \geq 0$ and $r > 0$ are design
parameters. We solve the control system~\eqref{eq:first-order} to find $x(t) =
e^{(p+\theta)t}$ and plug this into the performance index~\eqref{eq:perfind}
along with the form selected for the controller. Performing the integration over
time and letting $T \to \infty$, assuming that $p+\theta < 0$ then yields the
infinite-horizon optimal cost functional
\begin{equation} \mc{J}_\infty = -\frac{1}{4}\frac{q+r\theta^2}{p+\theta}.
\label{eq:inf-time-integrated-cost} \end{equation}
The optimal control parameter $\theta$ may be found as the appropriate root of
$\nabla_\theta \mc{J}_\infty$. 
\begin{align} 
    \begin{split} 
        \nabla_\theta \mc{J}_\infty &= -\frac{r}{4}\frac{(p+\theta)^2 - \left(p^2
        + \nicefrac{q}{r}\right)}{(p+\theta)^2} = 0, \\ 
        \therefore \theta^\star &= g(p) :=-p - \sqrt{p^2 + \nicefrac{q}{r}}, \\
        &\hspace{2mm} g^{-1}(\theta) = \frac{q}{2r\theta} - \frac{\theta}{2}.
    \end{split} 
    \label{eq:optimal_theta} 
\end{align}
The fact that $p \sim \mc{N}(\hat{p}, \sigma_p^2)$ implies that the optimal
control parameter has the probability density function
\begin{align*} 
    f_{\theta^\star}(\theta^\star) &= f_p\left(g^{-1}(\theta^\star)\right)
        \abs{\frac{d}{d\theta}g^{-1}(\theta^\star)}, \\ 
        &= \frac{1}{\sigma_p
        \sqrt{2\pi}}\left(\frac{1}{2}\left(1+\frac{q}{r{\theta^\star}^2}\right)\right)
        \exp{\left\{-\frac{1}{2\sigma_p^2}\left(
        \frac{q}{2r\theta^\star} - \frac{\theta^\star}{2} - \hat{p}
        \right)^2\right\}}, 
\end{align*}
where $f_p$ is the Gaussian probability density function with mean $\hat{p}$ and
variance $\sigma_p^2$. 

We can further eliminate the control parameter from the
expression for the optimal cost function $\mc{J}_\infty$ by
substituting for $\theta$ from
equation~\eqref{eq:optimal_theta}, yielding 
\begin{align*}
\mc{J}^\star = &h(p) := \frac{r}{2}\left( p + \sqrt{p^2 + \nicefrac{q}{r}} \right), \\
                &h^{-1}(\mc{J}^\star) = \frac{\mc{J}^\star}{r} -
                \frac{q}{4\mc{J}^\star}.
\end{align*}
Hence, the distribution of the optimal cost conditioned on the system parameter
$p$ is 
\begin{align*} 
    f_{\mc{J}^\star}(\mc{J}^\star) &=f_p\left(h^{-1}(\mc{J}^\star)\right)
        \abs{\frac{d}{d\theta}h^{-1}(\mc{J}^\star)}, \\
        &= \frac{1}{\sigma_p
        \sqrt{2\pi}}\left(\frac{1}{r} + \frac{q}{4{\mc{J}^\star}^2}\right)
        \exp{\left\{ -\frac{1}{2\sigma_p^2} \left(
        \frac{\mc{J}^\star}{r} - \frac{q}{4\mc{J}^\star} -
        \hat{p}\right)^2\right\}}.
\end{align*}
Notice that the distribution of both the optimal control parameter and the
optimal cost are elements of the exponential family that are not Gaussian. 

\begin{figure}[tb]
  \centering
  \includegraphics[width=0.55\linewidth]{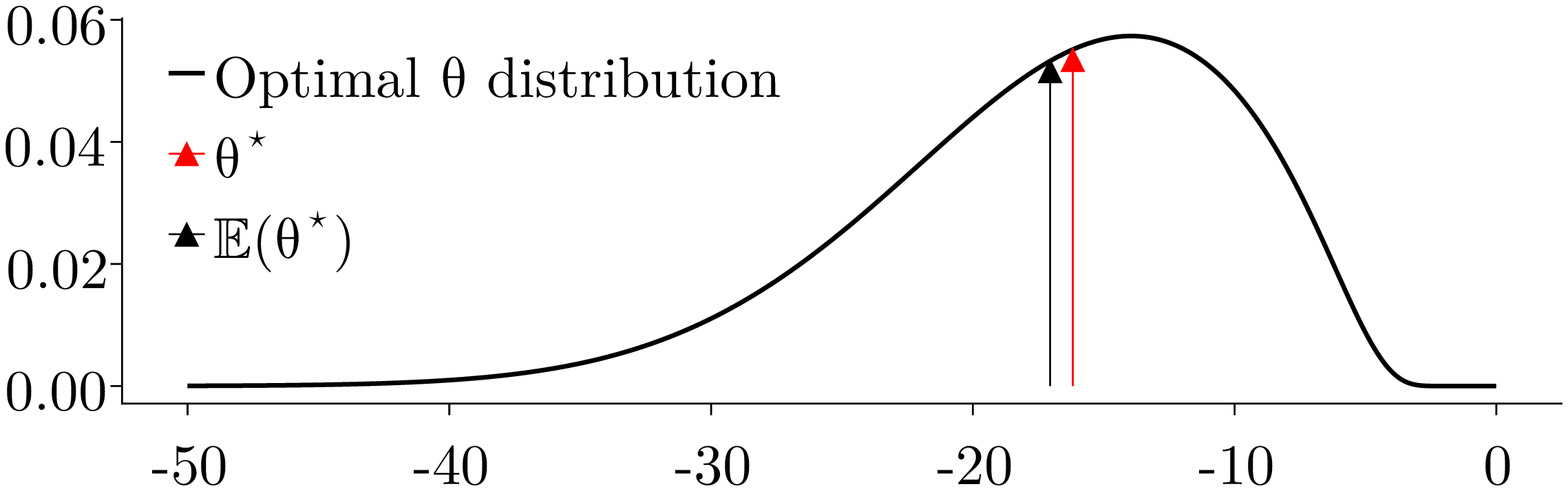}
  \caption{The optimal control parameter distribution given that the system
  parameter $p$ is normally distributed with mean $\hat{p} = 5$ and $\sigma_p
  = 5$. The red and black arrows respectively indicate the optimal control
  parameter without considering the randomness of $p$, and the expected value
  of the optimal control parameter distribution.}
  \label{fig:optimal_dist}
\end{figure}

%
In order to derive some quantitative results, let us assign some numerical
values to the parameters that define the optimal cost function $(q,r) = (100, 1)$, our best
guess $\hat{p} = 5$ of the system parameter $p$ and its standard deviation $\sigma_p = 5$.
The optimal control parameter and cost derived for this system whose model is
assumed to be known perfectly are given by $\hat{\theta}^\star = -16.180$ with
the corresponding estimated cost $\hat{\mc{J}}^\star = 8.090$.
This deterministic performance estimate turns out to be \textit{overconfident} when
uncertainties in the system parameter are present. 
For example, if the prior knowledge on the distribution of the system parameter
$p$ is utilized, the expected value of the controller parameter is found as
$\mathbb{E}[\theta^\star] = -17.046$ and the corresponding expected cost is
$\mathbb{E}[\mc{J}] = 8.523$. 
The controller from the deterministic training/optimization is not only
overconfident about its performance; but also is less robust against modeling
errors, as the Bayesian learning yields a closed-loop stable system for a wider
range of values of $p$.

Finally, Figure~\ref{fig:optimal_dist} shows the optimal control parameter
distribution given that the system parameter $p$ is normally distributed with
mean $\hat{p} = 5$, standard deviation $\sigma_p = 5$.
This figure also shows the mean values of the optimal control distribution with
the black arrow and the optimal control parameter a deterministic approach would
yield in red. 
We notice that the Bayesian learning that yields the optimal control parameter
distribution is more concerned about system stability due to the uncertainty in
the parameter $p$, a feat that the deterministic training may not reason about.



\subsection{Optimal Control under Parameter Uncertainty and Measurement
Noise}

Consider the scenario in which the system~\eqref{eq:first-order} is also subject
to measurement errors; that is, our measurement model for the state $x$ is
probabilistic and is distributed according to the Gaussian $\mc{N}(x,
\sigma^2)$. Since the controller uses this measurement to determine its action,
the closed-loop system has to be modelled as a stochastic differential equation
(SDE), given by
\begin{equation} 
    \begin{cases} 
    d x(t) = (p+\theta)x(t) \, d t + \theta\sigma \, d W_t, \\ 
    x(0) = 1,
    \end{cases} 
    \label{eq:first-order-SDE} 
\end{equation}
where $W$ denotes the Wiener process~\cite{evans2012introduction}. The initial
state is assumed deterministic and is set to unity for simplicity.  
%
The unique solution to this SDE is given by
\begin{equation}
    x(t) = e^{(p+\theta)t} + \theta\sigma \int_0^t e^{(p+\theta)(t-s)}dW_s.
    \label{eq:sol-sde}
\end{equation}
\vspace{-4mm}
\begin{lem}\label{lem:1}
    The conditional expectation $\mathbb{E}\left[\mc{J}\mid p\right]$ of the
    performance index~\eqref{eq:perfind} given the system parameter $p$ is 
    \begin{align*} 
      \mathbb{E}\left[\mc{J} \mid p\right] &=-\frac{1}{4}\frac{q+r\theta^2}{p+\theta}
       \left[ \theta^2 \sigma^2 T + \left(1 - e^{2T(p+\theta)}\right) \left(1 +
      \frac{1}{2}\frac{\theta^2\sigma^2}{p+\theta}\right)\right].  
    \end{align*} 
\end{lem}

\begin{pf}
Substituting the solution~\eqref{eq:sol-sde} of the
SDE~\eqref{eq:first-order-SDE} expression into the performance
measure~\eqref{eq:perfind} yields
\begin{align*} \mc{J} =
        & -\frac{1}{4}\frac{q+r\theta^2}{p+\theta}\left(1 +
        e^{2T(p+\theta)}\right) + 
        (q+r\theta^2)\theta\sigma \int_0^T
        e^{(p+\theta)t} \int_0^t e^{(p+\theta)(t-s)}dW_s
        dt \; + \\ & \hspace{0.5cm}\frac{1}{2}(q+r\theta^2)\theta^2
        \sigma^2 \int_0^T \left( \int_0^t
        e^{(p+\theta)(t-s)} dW_s  \right)^2 dt 
\end{align*}
The conditional expectation of this quantity given the system parameter $p$
under the distribution induced by the Wiener process may be computed in
closed-form using It\^{o} calculus:
\begin{align*} 
\mathbb{E}_W\left[\mc{J} \mid p \right]
        &= -\frac{1}{4}\frac{q+r\theta^2}{p+\theta}\left(1 -
        e^{2T(p+\theta)}\right) +
        (q+r\theta^2)\theta\sigma\int_0^Te^{(p+\theta)t}\cancelto{0}{\mathbb{E}_W\left[\int_0^t
        e^{(p+\theta)(t-s)} dW_s~\Big\rvert p\right]} dt + \\
        & \hspace{0.5cm} \frac{1}{2}(q+r\theta)^2\theta^2\sigma^2\int_0^T\mathbb{E}_W\left[\left(\int_0^t
        e^{(p+\theta)(t-s)} dW_s \right)^2~\bigg\rvert p\right] dt \\
        &=-\frac{1}{4}\frac{q+r\theta^2}{p+\theta}\left(1 -
        e^{2T(p+\theta)}\right) + 
        \frac{1}{2}(q+r\theta^2)\theta^2\sigma^2\int_0^T\left(\int_0^t
        e^{2(p+\theta)(t-s)} ds \right) dt \\ 
        &=-\frac{1}{4}\frac{q+r\theta^2}{p+\theta}\left(1 - e^{2T(p+\theta)}\right) + 
        \frac{1}{2}(q+r\theta^2)\theta^2\sigma^2\int_0^T
        -\frac{1}{2(p+\theta)}\left(1 - e^{2T(p+\theta)}\right) dt \\ 
        &=-\frac{1}{4}\frac{q+r\theta^2}{p+\theta} \left[ \theta^2 \sigma^2 T +
        \left(1 - e^{2T(p+\theta)}\right) \left(1 +
        \frac{1}{2}\frac{\theta^2\sigma^2}{p+\theta}\right)\right].
        & \pushright{\blacksquare}
\end{align*}
\end{pf}
It is easily shown that this quantity is positive for all $T>0$. Furthermore, it
blows up as the horizon $T$ is extended to infinity. This is not surprising
since a nonzero measurement noise causes the state to oscillate around the
origin, rather than asymptotically converging to it, incurring nonzero cost
all the while.


We have kept the system parameter $p$ constant in this analysis so far.
Uncertainty over this variable can be incorporated by taking a further
expectation \[ \mathbb{E}[\mc{J}] := \mathbb{E}_p\left[\mathbb{E}_W\left[\mc{J}
\mid p \right]\right], \] of $\mathbb{E}_W\left[\mc{J} \mid p \right]$ over $p$,
which must be accomplished numerically as it does not admit a closed-form
expression.  

\begin{figure}[tb]
  \centering
    \begin{minipage}{.48\textwidth}
        \centering
        \includegraphics[width=0.8\linewidth]{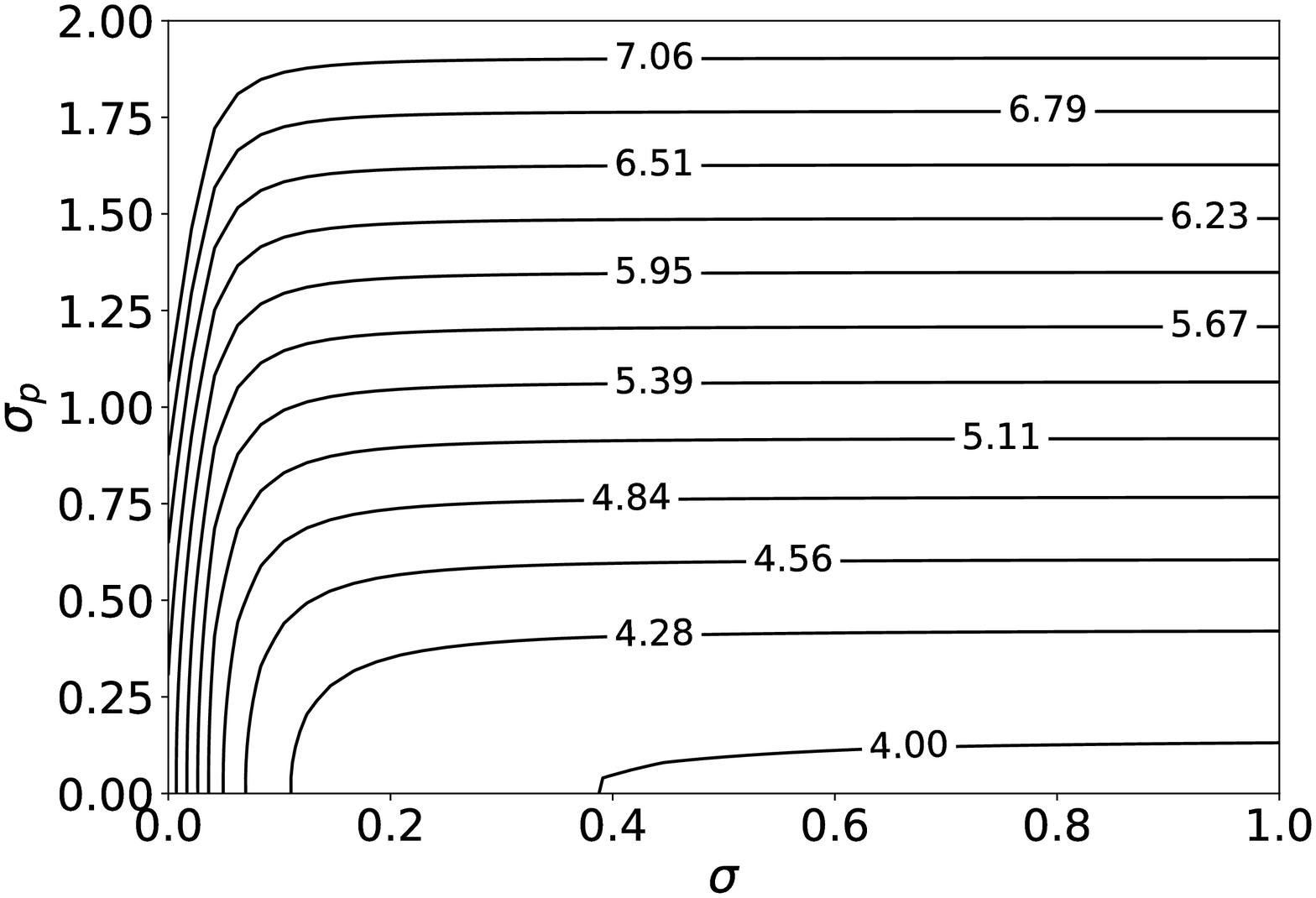}
        \caption{The optimal controller parameter magnitude $\abs{\theta^\star}$.}
        \label{fig:iwp-projections}
    \end{minipage}%
    \hfill
    \begin{minipage}{0.46\textwidth}
        \centering
        \includegraphics[width=0.8\linewidth]{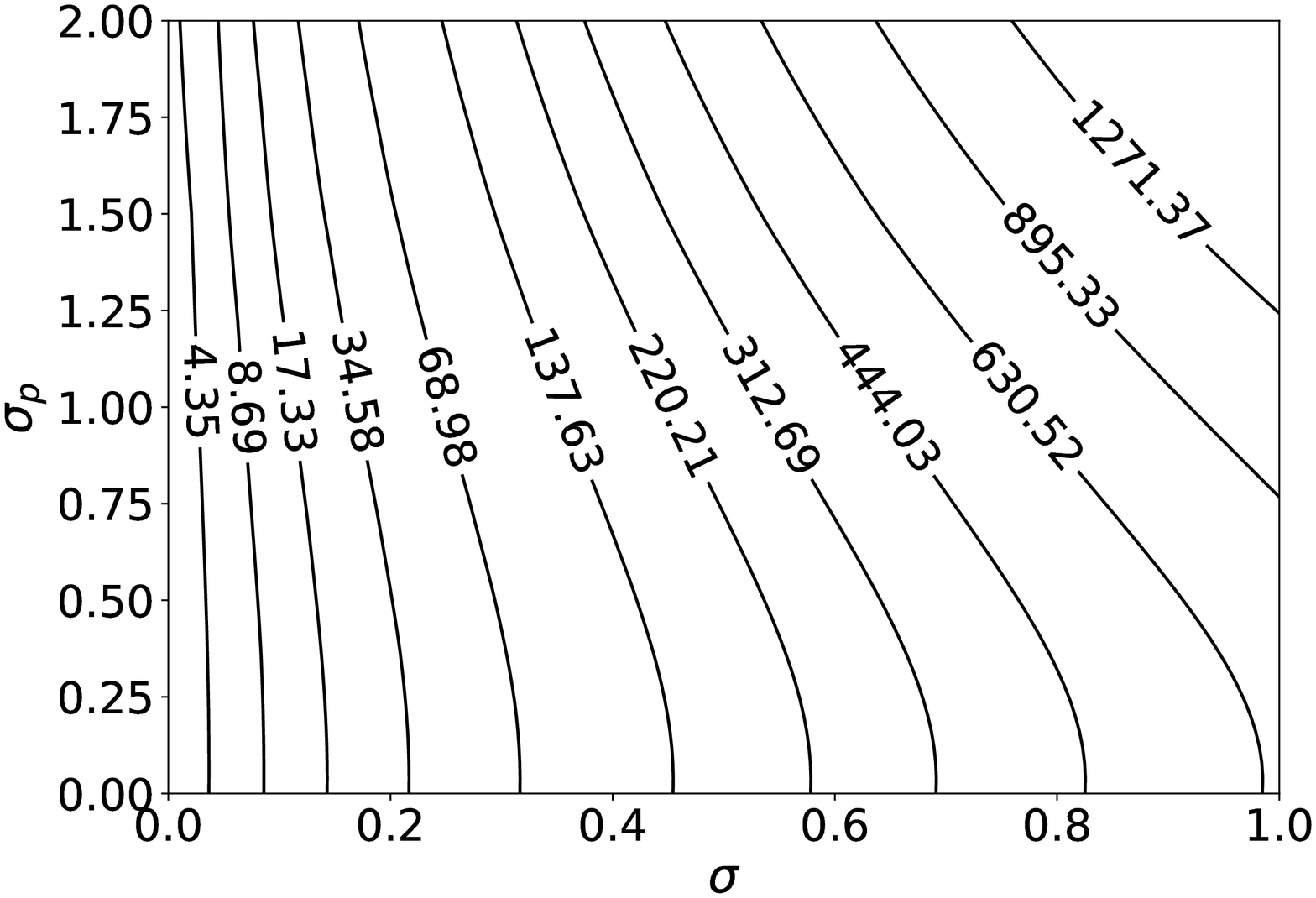}
        \caption{The minimal expected cost $\mathbb{E}[\mc{J}]$.}
        \label{fig:iwp-loss}
    \end{minipage}
  %
  \label{fig:optimal-ctrl-cost}
\end{figure}

We can then minimize $\mathbb{E}[\mc{J}]$ over the control parameter in order to
study the effects of both kinds of uncertainties on the optimal controller.
Such a study is provided in Figures~\ref{fig:iwp-projections}
and~\ref{fig:iwp-loss}, where we have plotted the optimal control parameter
$\theta^\star$ and the minimal expected cost $\mathbb{E}[\mc{J}]$ as a function
of the standard deviations of the measurement noise $\sigma$ and the system
parameter $\sigma_p$. The constants we used to generate the data are given by
$q=r=1$ and $T=\hat{p}=3$. Our first observation is that the magnitude of the
optimal control parameter is an increasing function of system parameter
uncertainty and a decreasing function of measurement uncertainty. Our second
observation is that if the measurement noise is small, then the optimal control
parameter is insensitive to system parameter uncertainty as long as this
uncertainty is small. The optimal cost shares this insensitivity for an even
wider range of values of $\sigma_p$. In a similar vein, if the uncertainty in
the system parameter is large, then the optimal control parameter is insensitive
to the magnitude of the measurement noise. However, the optimal cost is still
sensitive to this quantity.

\paragraph{Key Takeaways}
There are several advantages, to be inferred from the analyses in this section,
of employing Bayesian learning to find the optimal control parameter $\theta$.
Note that the optimal control parameter and cost derived for this system whose
model is assumed to be known perfectly without measurement noise are given by
$\theta_d^\star = -6.162$ and $\mc{J}_d^\star := \mc{J}(\theta_d^\star) =
3.081$.  
This deterministic performance estimate is greatly overconfident when
uncertainties in the system parameter and measurement are present. For example,
if $\sigma = \nicefrac{1}{10}$ and $\sigma_p = \nicefrac{\hat{p}}{5} =
\nicefrac{6}{10}$ then the expected cost with this controller parameter is, in
fact, $\mathbb{E}[\mc{J}] = 14.907$ and the overconfidence is an increasing
function of both kinds of uncertainties.

The deterministic optimal controller certainly yields a controller parameter
$\theta_d^\star$ that yields a stable system for a range of values for the
system parameter $p$. 
In the numerical example, even if our best belief $\hat{p}$ of $p$ is erroneous
by $100\%$, this controller will stabilize the system assuming no measurement
noise. 
\begin{wrapfigure}{r}{0.65\textwidth}
    \vspace{-1mm}
    \centering
    \includegraphics[width=0.65\textwidth]{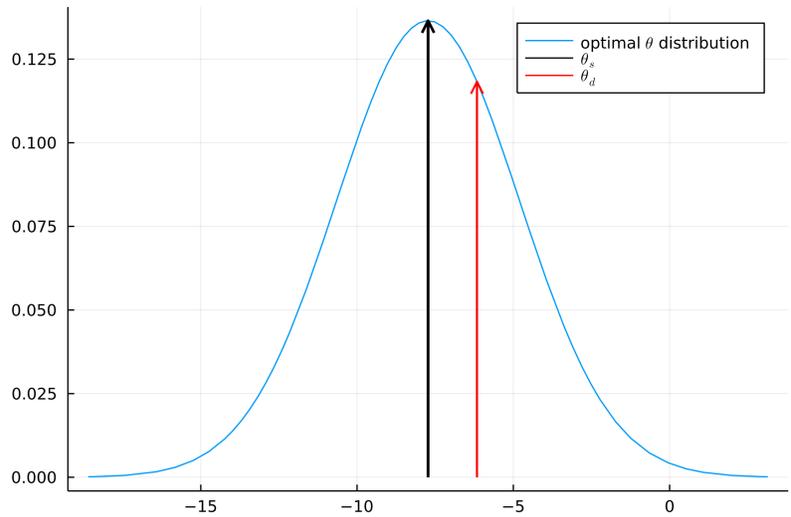}
    \caption{Laplace (Gaussian) approximation of the optimal control parameter
    distribution (shown in \textcolor{blue}{blue}) given that the system
    parameter $p$ is distributed according to a truncated normal with mean
    $\hat{p} = 3$ and $\sigma_p = 2$ and bounds $(0, 6)$. The deterministic
    optimal controller $\theta^\star_d$ is shown in black and the Bayesian
    solution $\theta^\star_s$ is shown in \textcolor{red}{red}.}
    \label{fig:optimal_dist_full}
    \vspace{-4mm}
\end{wrapfigure}
However, consider the situation where we are pretty certain about the
system parameter. Then, in the presence of measurement noise, this controller
parameter induces a much larger expected cost than the optimal controller
parameter, $\theta_s^\star$ that Bayesian learning yields (which has smaller
magnitude, i.e., $\abs{\theta_s^\star} < \abs{\theta_d^\star}$). On the flip
side, if the system parameter uncertainty is large, (e.g. $\sigma_p >
\nicefrac{\hat{p}}{2}$), then, again, Bayesian learning is able to account for
this fact to yield a controller parameter that is more robust. In this case, the
situation is reversed and we find out $\abs{\theta_d^\star} <
\abs{\theta_s^\star}$. Clearly, $\theta_s^\star$ stabilizes the system for a
wider range of the true values of the system parameter than $\theta_d^\star$
does. When both measurement and system parameter uncertainties are present,
Bayesian learning is able to precisely strike the trade-off between these
competing uncertainties and yield controller parameter that results in a much
better expected cost than a deterministic optimization does.

Finally, Figure~\ref{fig:optimal_dist_full} shows the Laplace (Gaussian)
approximation to the optimal control parameter distribution given that the
measurement noise has standard deviation $\sigma = \nicefrac{1}{10}$, and the
system parameter $p$ is distributed according to a truncated normal with mean 
$\hat{p} = 3$, standard deviation $\sigma_p = 2$ and lower and upper bounds $(0,
6)$. This figure also shows the mean values of the optimal control distribution
with the black arrow and the optimal control parameter a deterministic approach
would yield in red. We notice that the Bayesian learning that yields the
optimal control parameter distribution is more concerned about system stability
due to the uncertainty in the parameter $p$, a feat that the deterministic
training may not reason about.

%% file: contents/conclusion.tex
\section{Conclusion}
\label{sec:conclusion}

We solve the optimal control problem under the deterministic and Bayesian
settings.
%
%
The optimal control parameter changes sharply as a function of uncertainties in
system parameters and measurements. 
This encourages the use of the Bayesian approach, which takes this variation
into account, coming up with controllers that yield closed-loop stable systems
for a wider range of system parameters.
These findings motivate the use of Bayesian solutions to the control search
problems, paving the way for the integration of Bayesian learning in data-driven
control design techniques. 